# Multimodal Sparse Classifier for Adolescent Brain Age Prediction

Peyman Hosseinzadeh Kassani, Alexej Gossmann, and Yu-Ping Wang, *Senior Memb*er, *IEEE*

*Abstract*— The study of healthy brain development helps to better understand the brain transformation and brain connectivity patterns which happen during childhood to adulthood. This study presents a sparse machine learning solution across whole-brain functional connectivity (FC) measures of three sets of data, derived from resting state functional magnetic resonance imaging (rs-fMRI) and task fMRI data, including a working memory n-back task (nb-fMRI) and an emotion identification task (em-fMRI). These multi-modal image data are collected on a sample of adolescents from the Philadelphia Neurodevelopmental Cohort (PNC) for the prediction of brain ages. Due to extremely large variable-to-instance ratio of PNC data, a high dimensional matrix with several irrelevant and highly correlated features is generated and hence a pattern learning approach is necessary to extract significant features. We propose a sparse learner based on the residual errors along the estimation of an inverse problem for the extreme learning machine (ELM) neural network. The purpose of the approach is to overcome the overlearning problem through pruning of several redundant features and their corresponding output weights. The proposed multimodal sparse ELM classifier based on residual errors (RES-ELM) is highly competitive in terms of the classification accuracy compared to its counterparts such as conventional ELM, and sparse Bayesian learning ELM.

*Index Terms*— Brain age prediction, Extreme learning machine, Inverse problem, Neuroimaging, Sparsity

## I. INTRODUCTION

ONE of the most important periods for neurodevelopmental changes in the brain is during adolescence. The variations in the brain of an adolescent over this critical period may even determine the adult functioning and mental health for the rest of life [11]. Analyzing human brain anatomy during this critical period is very useful for neuroscientists to gain insights into tasks such as measurement of structural neurodevelopment, brain maturation, adults' cognitive superiority and brain regions involved therein [11]. This fundamental period has motivated neuroscientists to research brain aging for a better understanding of the brain transformation and its connectivity patterns. Studies of brain age prediction may raise some important questions including: how to discriminate between a typical brain development and a neuropsychiatric disorder; why some brain regions seem functionally connected based on the observed neural activity while other brain regions are less interdependent, and more interestingly to us, how different imaging modalities can be used to study those brain regions.

Blood oxygen level dependent functional magnetic resonance imaging (BOLD fMRI, or simply fMRI) is a noninvasive radiological examination which, among many other applications, can help to understand neurodevelopmental processes of the human brain in the transition to adulthood [5]. Functional MRI measures the neural activity in brain voxels over time, and therefore it can be acquired while the subject performs a cognitive task or when the subject is idle, in order to observe changes in neural activity corresponding to the task on and off conditions or during rest. The PNC data to be analyzed in Section V, includes three fMRI modalities, such as nb-fMRI, which is a standard working memory task, em-fMRI, and rs-fMRI [30]. Moreover, each fMRI modality can be used to obtain measures of association between brain regions or voxels which is known as functional connectivity (FC). One way to obtain FC measures is to look at the pair-wise correlation coefficients between the time series of different brain regions. The fMRI data as well as the derived FC data are high-dimensional and usually there are many more variables than the number of individuals, e.g., a single MRI may consist of tens of thousands of voxels, many of which are highly correlated and hence difficult to deal with [12].

While neural activity and connectivity patterns are highly complex and non-interpretable even for domain expert, machine learning helps to discover useful knowledge hidden in the data [16, 17]. In this study, we propose a new machine learning algorithm based on a sparse learning method embedded into the extreme learning machine (ELM) framework [14, 15], which is a single hidden layer feedforward neural network (SLFN) classifier. We apply this sparse ELM to three modalities of fMRI data for the task of brain age prediction.

The rest of this paper is organized as follows: in Section II we discuss preliminaries, and in Section III some of the related works as well as the motivation and contributions of this study are explained. The proposed residual error based sparse ELM (RES-ELM) method for feature selection is explained in

"This work was partially supported by NIH (R01GM109068, R01MH104680, R01MH107354, P20GM103472, R01EB020407, 1R01EB006841) and NSF (#1539067)."
(Corresponding author: Yu Ping Wang)

P. H. Kassani and Y.-P. Wang are with the Department of Biomedical Engineering, Tulane University, New Orleans, LA, 70118 USA (e-mails: peymanhk@tulane.edu , wyp@tulane.edu).
A. Gossmann is with the Bioinnovation Ph.D. Program, Tulane University, New Orleans, LA 70118 USA (e-mail: agossman@tulane.edu).

Section IV. Experiments on PNC data are described in Section V. Concluding remarks are given in Section VI.

## II. PRELIMINARIES

ELM algorithms, originally proposed by Huang et al. [14], are single layer feedforward neural networks (SLFNs) [9, 10, 15]. In ELM there is no need to learn weights from the input layer to the hidden layer since all weights are set randomly. Let $N$ be the number of training instances and $p$ be the dimension of the feature vector in the training dataset $D = (\mathbf{x}_i, t_i)$, $i = 1, 2, \ldots, N$, where $\mathbf{x}_i \in \mathbb{R}^{p \times 1}$ denotes the input vector and $\mathbf{t}_i \in \mathbb{R}^{c \times 1}$ is the target vector with $c$ classes. To randomly map the $p$-dimensional feature vector in the input space to $L$ dimensional basis vector where $L > p$, a function $K(\mathbf{a}, b, \mathbf{x})$ is required where $\mathbf{a} \in \mathbb{R}^{L \times 1}$ and $b$ are random weights and bias respectively. The purpose is to find the best random subspace wherein data can be more accurately regressed. The approximation of the target values $\mathbf{t}_j$ of $N$ data instances with $L$ hidden neurons depends on the activation function $K$ and its parameters.

Let $\mathbf{T} \in \mathbb{R}^{N \times c}$ be a matrix with its rows given by $k_L(\mathbf{x}_j)$ $j = 1, 2, \ldots, N$, $\mathbf{B} \in \mathbb{R}^{L \times c}$ be a matrix with rows given by $\boldsymbol{\beta}_i$ for $i = 1, 2, \ldots, L$, and $\mathbf{H} \in \mathbb{R}^{N \times L}$ be a matrix where its (i,j)th element is given by $K(\mathbf{a}_i, b_i, \mathbf{x}_j)$, $i = 1, 2, \ldots, L, j = 1, 2, \ldots, N$, then,

$$\mathbf{HB} = \mathbf{T} \quad (1)$$

Note that in the matrix form $\mathbf{H} \in \mathbb{R}^{N \times L}$ is termed as the *design matrix* and $\mathbf{T} \in \mathbb{R}^{N \times c}$ is the target matrix, and $\mathbf{B} \in \mathbb{R}^{L \times c}$ is the group of output weights which can be estimated in least-square sense as follows:

$$\mathbf{B} = \mathbf{H}^\dagger \mathbf{T} = (\mathbf{H}^T \mathbf{H})^{-1} \mathbf{H}^T \mathbf{T} \quad (2)$$

where $\mathbf{H}^\dagger$ is the left Moore-Penrose generalized inverse of $\mathbf{H}$. Since the number of features is much higher than the number of subjects, i.e. $L > N$, the right Moore-Penrose generalized inverse of $\mathbf{H}$, which is $\mathbf{H}^\dagger = \mathbf{H}^T (\mathbf{HH}^T)^{-1}$, is used. Given $\mathbf{B}$, one can simply obtain the prediction output $\hat{\mathbf{T}} = \mathbf{HB}$.

In next section we discuss several works conducted on the brain age prediction problem.

## III. RELATED WORKS

Several studies have been conducted on brain age prediction over the last decade [1-8]. The motivation is to understand the brain transformation and its connectivity patterns during childhood to adulthood. We first review studies conducted on unimodal data, and then we describe the recent work on multimodal data.

In [1], rs-fMRI data from 50 preterm-born infants (postmenstrual age) and 50 term-born control infants studied within the first week of life were used for the prediction of brain maturity. Using 214 regions of interest (ROI), binary support vector machines (SVM) can correctly classify 84% of them. Inter- and intra-hemispheric connections of the brain were important for this high classification rate, indicating that widespread changes in the brain's functional network architecture associated with preterm birth are detectable by term equivalent age.

In [3], a hidden Markov model was devised for MRI image structures. The root mean squared error (RMSE) and the mean absolute error (MAE) were both used to evaluate the accuracy of the age prediction. The proposed approach has been able to effectively perform the prediction by using much smaller training samples. The results have also shown its superior brain age prediction ability in comparison to other methods for brain age prediction, including relevance vector machine for regression (RVR), and quantitative brain water maps (BWM), with both RMSE and MAE metrics.

Another interesting study [6] improves the brain-age prediction accuracy by optimizing resampling parameters using Bayesian optimization. This study trains support vector machines (SVM) on 2003 healthy individuals (aged 16–90 years) for two purposes: (i) distinguish between young (<22 years) and old (>50 years) brains (classification problem) and (ii) predict chronological age. This study also uses Bayesian optimization to derive case-specific pre-processing parameters and to identify optimal voxel size of brain and smoothing kernel size of SVM for each task. Adjusting parameters is performed by adaptively sampling the parameter space across a range of possible parameters. When distinguishing between young and old brains, a classification accuracy of 88.1% was achieved and for predicting chronological age, a MAE of 5.08 years was achieved.

Study [8] lists several research papers assessing brain age in neurological and psychiatric diseases. This qualitative work also brings up controversies surrounding brain age and highlights emerging trends such as the use of multimodality neuroimaging and the employment of 'deep learning' methods.

Study [7] investigates how multimodal brain-imaging data improves age prediction. Five sources of neuroimaging data entered the age prediction models. Two sources represent brain connectivity in different spatial resolutions and three sources originate from brain anatomy. After extracting feature vectors for each subject and modality, they were stacked together for the age prediction analysis. First, linear support vector regression models (SVR) were used to predict age from neuroimaging data (single-source models). Next, predictions from the single-source models were stacked with random forest (RF) regression models. It is found that multimodal data improves brain-based age prediction.

### A. Motivation and contributions

From above reviews, the studies of [2, 4, 7] used multimodal data for brain age prediction to gain better performance. The necessity of the creation of a sparse classifier which can detect important features on multiple modalities is demonstrated in the mentioned studies but lacks of an effective model. This motivated us to devise a new model, which uses multimodality and sparse feature learning jointly for the problem of the brain age prediction. This study has the following contributions:

1- A new sparse model takes advantage of the computation of inversion in least squares error (LSE) minimization based neural networks (NNs) and reduces the errors of this computation. The proposed model is applied to hidden neurons of ELM. This way of sparsity can be applied to every kind of

LSE based NNs.

2- To validate the reliability of the proposed RES-ELM, real experiments are conducted on three sets of fMRI modalities of brain imaging; namely nb-fMRI, rs-fMRI, and em-fMRI. The experiments show that multimodal approaches achieve better accuracy than unimodal ones.

## IV. THE RESIDUAL ERROR BASED SPARSE ELM

### A. Overview

The proposed RES-ELM includes three sequential steps as shown in Fig. 1. The first step applies group-level independent component analysis (group ICA) [20, 21] to three sets of fMRI modalities to reduce the dimension of data and computes each subject's FC based on the ICA time courses. The second step sparsifies the neural network with the idea of ranking residual errors. To do so, the residual error of $\mathbf{H}^\dagger \mathbf{H}$ for hidden neurons ranking is computed. This way we are able to *rank hidden neurons* using the residual error values obtained for each neuron (Section IV.C.1). The third step prunes hidden neurons from the network and dimensions of data from $\mathbf{H}$ (Section IV.C.2).

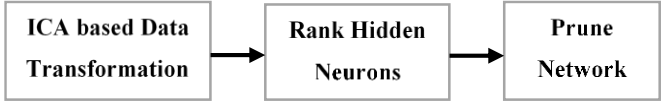

Fig. 1. The flowchart for the proposed RES-ELM algorithm.

### B. Dimension reduction with ICA

ICA is able to unmix unknown signal sources to a new set of signals [22, 23]. ICA uses a more restrictive and stronger constraint compared with principal component analysis (PCA) wherein different dimensions are assumed to be uncorrelated [23]. ICA has been successfully applied to several types of medical data including neuroimaging [25-28].

### C. Hidden neuron ranking via residual error

Referring to the second step in Fig. 1, our proposed method based on the residual error for pruning the columns of the design matrix $\mathbf{H}$ of ELM is explained. In the next section we explain where the residual errors occur in LSE minimization based NNs.

*1) Truncation errors in singular value decomposition*

When $L > N$, direct computation of $\mathbf{H}^\dagger = \mathbf{H}^T(\mathbf{HH}^T)^{-1}$ for solving $\mathbf{B} = \mathbf{H}^\dagger \mathbf{T}$ in (2), i.e., $\mathbf{B} = \mathbf{H}^\dagger \mathbf{T} = \mathbf{H}^T(\mathbf{HH}^T)^{-1}\mathbf{T}$ causes numerical instability [18]. A remedy to overcome this instability is to employ Singular Value Decomposition (SVD) which is more robust to the numerical errors [19]. SVD decomposes matrix $\mathbf{H}$ as follows,

$$\mathbf{H} = \mathbf{PRQ}^T \quad (3)$$

where $\mathbf{P}$ is an $N \times N$ orthogonal matrix with column vectors as eigenvectors of $\mathbf{HH}^T$, $\mathbf{Q}$ is also an $L \times L$ orthogonal matrix with column vectors as eigenvectors of $\mathbf{H}^T\mathbf{H}$, and $\mathbf{R}$ is a truncated diagonal matrix of size $N \times L$ with diagonal elements called singular values. With this setting, $\mathbf{B}$ can be reformulated as:

$$\mathbf{B} = \mathbf{H}^\dagger \mathbf{T} \approx \mathbf{Q}\mathbf{R}^\dagger \mathbf{P}^T \mathbf{T} \quad (4)$$

where

$$\mathbf{R}^\dagger = \begin{cases} \frac{1}{\delta_k} & \text{if } \delta_k > \varepsilon \\ 0 & \text{if } \delta_k \leq \varepsilon \end{cases} \quad (5)$$

where $\varepsilon$ is a threshold value and $\delta_k$ is the $k^{\text{th}}$ singular value of $\mathbf{R}$. There is an issue with this setting of SVD. In practice, the very small singular values are usually set to zero. The reason is to avoid inflation of these small values in the computation of the inversion in (4). The zeroing of singular values poorly provokes numerical errors [18]. In this study, this numerical error is named *residual error*. We make use of these small-scale error values and design a sparse learner which can greatly improve the final classification accuracy. The residual error has a direct role in the computation of $\mathbf{B} = \mathbf{H}^\dagger \mathbf{T}$ since it jeopardizes the estimation of the pseudoinverse of $\mathbf{H}$. In the next section we explain how this idea is used for pruning hidden neurons of ELM neural network.

*2) Hidden neurons pruning*

As discussed above, one can take advantage of the errors that occur in the computation of $\mathbf{B}$ and design a new model to prune the neural network. To design such a pruning tool, we should carefully look at $\mathbf{H}^\dagger$ where errors occur. In fact, we know that if $\mathbf{H}$ has linearly independent columns then $\mathbf{H}^\dagger \mathbf{H} = \mathbf{I}$ where $\mathbf{I}$ is the identity matrix. In practice we have $\mathbf{H}^\dagger \mathbf{H} \approx \mathbf{I}$. So due to the property $\mathbf{H}^\dagger \mathbf{H} \approx \mathbf{I}$, we name $\mathbf{H}^\dagger \mathbf{H}$ the *pseudo* identity matrix and denote it by $\hat{\mathbf{I}} \in \mathbb{R}^{L \times L}$. The diagonal and off-diagonal elements of $\hat{\mathbf{I}}$ slightly differ from one and zero respectively, because of the zeroing of negligible and small-scale singular values in (5).

The square matrix $\hat{\mathbf{I}} \in \mathbb{R}^{L \times L}$ has $L$ columns/rows, equal to the number of hidden neurons in the ELM neural network. This helps us to identify and prune those hidden neurons that cause large residual errors. We only need to know how much $\mathbf{I} \in \mathbb{R}^{L \times L}$ deviates from $\hat{\mathbf{I}} \in \mathbb{R}^{L \times L}$ for every column/row. The procedure will be elaborated in the next section.

As an evidence for the observation of the residual errors, the deviation of $\mathbf{I} \in \mathbb{R}^{L \times L}$ from $\hat{\mathbf{I}} \in \mathbb{R}^{L \times L}$ is illustrated using 511 instances from the 'rs-fMRI' PNC dataset.

In Fig. 2, we show the element-wise subtraction between matrices $\mathbf{H}^\dagger \mathbf{H}$ and identity matrix $\mathbf{I}$ using ELM with Sigmoid hidden nodes. As the number of hidden neurons, $L$ increases, the absolute value of the subtraction between $\mathbf{H}^\dagger \mathbf{H}$ and $\mathbf{I}$ monotonically increase as well. This demonstrates the extent to which the computation of $\mathbf{H}^\dagger$ suffers from the truncation error as in (5).

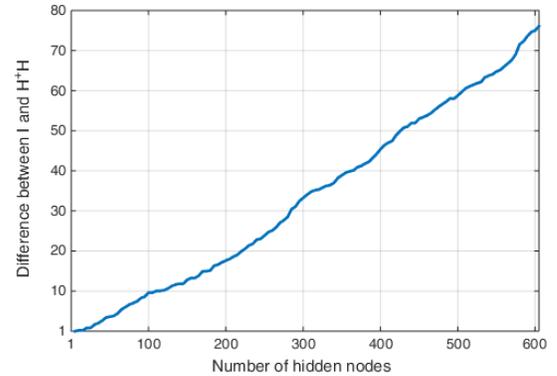

Fig. 2. Observation of residual errors for rs-fMRI data

*Lemma 1.* Let $\mathbf{H}_{L-1} = [K(\mathbf{a}_i, b_i, \mathbf{x}_j)]$, $i = 1, \ldots, N, j = 1, \ldots, L$ be the positive definite design matrix of ELM with $L$ hidden neurons and $N$ instances. When a hidden neuron is added, $\mathbf{H}_{L-1}$ is updated to $\mathbf{H}_L = [\mathbf{H}_{L-1} \ \mathbf{h}_L] = [K(\mathbf{a}_i, b_i, \mathbf{x}_j)], (i = 1,2, \ldots, N, j = 1,2, \ldots, L+1)$. Following (5), the residual error of $\mathbf{H}_{L-1}$ can be defined by

$$\mathrm{E}(\mathbf{H}_{L-1}) = \sum_{i:(\sqrt{\mu_i}-\varepsilon)\le 0} \sqrt{\frac{1}{\mu_i}} \quad (6)$$

where $\varepsilon$ is the threshold value and the eigenvalue spectrum for $\mathbf{H}_{L-1}$ is $\{\mu_i | i = 1,2, \ldots L\}$ with $L$ hidden neurons. Likewise, the residual error of $\mathbf{H}_L$ with $L+1$ hidden neurons is given by $\mathrm{E}(\mathbf{H}_L) = \sum_{i:(\sqrt{\gamma_i}-\varepsilon)\le 0} \sqrt{\frac{1}{\gamma_i}}$, where $\{\gamma_i | i = 1, \ldots, L+1\}$ is the corresponding eigenvalue spectrum. Note that $\mu_i$ and $\gamma_i$ are the eigenvalues of $\mathbf{H}_{L-1}^T \mathbf{H}_{L-1}$ and $\mathbf{H}_L^T \mathbf{H}_L$ respectively. It can be shown that:

$$\mathrm{E}(\mathbf{H}_L) \ge \mathrm{E}(\mathbf{H}_{L-1}) \quad (7)$$

The proof can be found in Appendix section. Referring to the Lemma 1 (and its proof), an important outcome is that the growth of errors in the current ELM network $\mathrm{E}(\mathbf{H}_L)$ depends on the new column $\mathbf{h}_L$ added to $\mathbf{H}_L$. The construction of $\mathbf{h}_L$ is by random generation of the input weight $\mathbf{a}$ and bias $b$ and the type of activation function. Since $\mathbf{a}$ and $b$ are randomly generated, there is no control on singular values in (5) and on residual errors as well. So, we desire to prune those neurons that have highest values of residual errors.

We are also interested in the relative difference between $\mathbf{H}^\dagger \mathbf{H}$ and $\mathbf{I}$. Fig. 3 shows this difference is significant and is nearly 40% for 600 hidden neurons, which necessitates to prune some neurons resulting in high errors within the ELM neural network.

To compute the relative difference we first do element-wise subtraction between matrices $\mathbf{H}^\dagger \mathbf{H}$ and $\mathbf{I}$ and then the sum of the absolute value of this subtraction is divided to the sum of elements of the identity matrix $\mathbf{I}$ for every iteration. As it can be seen some neurons help to decrease the relative error while some highly increase the relative error. We use this relative difference as a criterion for pruning hidden neurons of ELM.

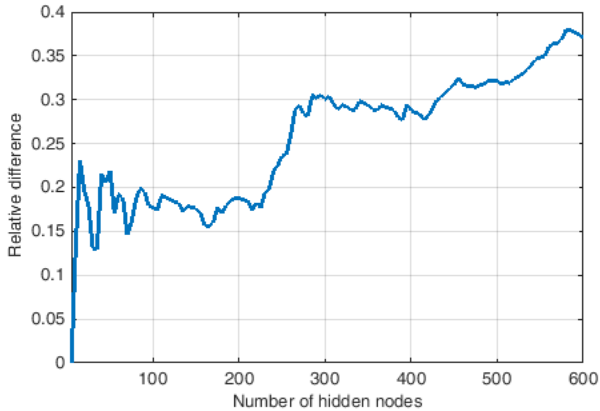

Fig. 3. Relative difference between $\mathbf{H}^\dagger \mathbf{H}$ and $\mathbf{I}$ for rs-fMRI data

*D. ELM sparsification stage*

In this section we show how residual errors can be used to make the design matrix $\mathbf{H}$ of ELM sparse. This is performed using a feedforward feature selection strategy. Based on the definition of the pseudo identity matrix $\hat{\mathbf{I}} \in \mathbb{R}^{L \times L}$, we design *Algorithm 1* for pruning ELM using residual errors called RES-ELM.

We first create the design matrix $\mathbf{H}$ with $L_{\max}$ hidden neurons using input weights $\mathbf{a}$ and bias $b$. We then choose a single hidden neuron and the column from $\mathbf{H}$ and incrementally add more hidden neurons as candidate neurons. The purpose is to minimize the relative error which is obtained for neuron $j$ as follows:

$$v_j = \frac{\|\hat{\mathbf{I}}_j - \mathbf{I}_j\|_1}{sum(\mathbf{I})} \quad (8)$$

where $\mathbf{I}_j \in \mathbb{R}^L$ is $j^{\text{th}}$ column vector of the identity matrix $\mathbf{I}$ and $\hat{\mathbf{I}}_j \in \mathbb{R}^L$ is $j^{\text{th}}$ column vector of $\hat{\mathbf{I}}$ and $sum(\mathbf{I})$ adds all the element of the matrix $\mathbf{I}$. The $v_j$ stores the relative error for neuron $j$. Equation (8) is computed for all $L_{\max}$ neurons. Among all neurons, the one which has the lowest relative error $v_j$ is kept and the index of the candidate neuron $\varphi_j$ is stored at the vector $\boldsymbol{\varphi}$.

So, we are left with $L_{\max} - 1$ hidden neurons as candidates. The relative error for remaining neurons is found through (8). And, the index and value of residual error of next candidate neuron are added to $\boldsymbol{\varphi}$ and $\mathbf{v}$ respectively. This procedure is continued until we reach the maximum number of neurons we would like to keep, which is $S_p\%$ of $L_{\max}$. This parameter $S_p$ is called sparsity parameter and is smaller than $L_{\max}$.

With update of $\boldsymbol{\varphi}$ and $\mathbf{v}$ we can observe which column vectors of $\mathbf{H}$ have higher relative error values and also, we can easily find their index values which should be selected from $\mathbf{H}$ and other columns should be pruned. What remains after pruning is a sparse design matrix $\mathbf{H}_{sp} \in \mathbb{R}^{N \times l}$ where $l \ll L_{max}$ is the number of column vectors in $\mathbf{H}$.

The RES-ELM is summarized in *Algorithm 1* as follows:

---
ALGORITHM 1. RES-ELM

**Input:**
1) The maximum number of Hidden neurons $L_{\max}$ and sparsity parameters $S_p \epsilon \ (1, 2, \ldots, 30)$ and $\boldsymbol{\varphi} = \{\ \}$
2) Randomly generate weights $\mathbf{a}$ and $b$ and matrix $\mathbf{H}$.
**Body:** (*%% Residual error-based pruning*)
**For** every $S_p$
 **While** size($\boldsymbol{\varphi}$) = Sp% * $L_{max}$
  **For** $j = 1$ to $L_{max}$
   3) Compute $\mathbf{H}^\dagger \mathbf{H} = \hat{\mathbf{I}}$.
   4) Compute $v_j = \frac{\|\hat{\mathbf{I}}_j - \mathbf{I}_j\|_1}{sum(\mathbf{I})}$ and store $v_j$ in *temp*.
   5) Find minimum $v_j$ via *temp*, store it in $\mathbf{v}$ and its index $\varphi_j$ in $\boldsymbol{\varphi}$.
   6) Remove index $\varphi_j$ from the pool of neurons and $L_{max} = L_{max}$ -1.
  **End For**
 **End While**
**Output:**
 **For** every $z$
   7) Compute $\mathbf{B}_{sp} = (\mathbf{H}_{sp}^T \mathbf{H}_{sp} + z\mathbf{I})^{-1} \mathbf{H}_{sp}^T \mathbf{T}$
   8) Compute the network output using $\hat{\mathbf{T}} = \mathbf{H}_{sp} \mathbf{B}_{sp}$.
 **End For.**
**End For**

---

In *Algorithm 1*, and step 7, the term $\mathbf{H}_{Sp}$ refers to the matrix

**H** after the pruning. Note that if $S_p$ is set to 100, the RES-ELM degenerates to conventional ELM. Next section discusses more about the methods.

## V. Experiments

In this section, we explain the way for data preparation and also we evaluate the proposed model in terms of feature selection and classification accuracy on three sets of imaging modalities.

### A. Data preparation and experiment setup

The data used in this study originates from the Philadelphia Neurodevelopmental Cohort [30] (PNC), which is a collaborative research effort between the Brain Behavior Laboratory at the University of Pennsylvania and the Center for Applied Genomics at the Children's Hospital of Philadelphia [30].

Nearly 900 adolescents with the ages in the range from 8 to 21 years old are chosen for our experiment. We preprocess data using standard brain imaging techniques through SPM12. They include motion correction, spatial normalization to standard MNI space (spatial resolution $2 \times 2 \times 2$mm) and spatial and temporal smoothing with a 3mm FWHM Gaussian kernel. The data preprocessing is borrowed from [29].

Some notes for the experimental setup:
1) Following the pre-processing steps of Section 6.1, the pre-processed fMRI data were decomposed into spatial maps and subject-specific time courses using a group-level spatial ICA [1] as implemented in the group ICA of fMRI toolbox (GIFT) [20]. Group ICA was applied to all three fMRI modalities, whereby the number of ICA components was set to C = 100. The C subject-specific time courses were then used to estimate a functional connectivity profile, or functional connectome (FC), for each fMRI modality for each subject. Generally, a subject's FC can be obtained by calculating Pearson correlations between the blood-oxygenation-level dependent (BOLD) time series corresponding to each pair of voxels or regions of interest (ROI) in the subject's fMRI image. However, in this work we chose to compute the FC on group ICA time courses rather than on the BOLD fMRI time series directly, in order to consolidate the relevant signals within a smaller number of temporal components which are relatively more informative. Thus, each subject's FC was estimated from the C subject specific time courses as their $C \times C$ sample covariance matrix. For further analysis each subject's estimated FC was flattened into a vector of length $\frac{C(C-1)}{2} = 4950$. Finally, the $n = 845$ (n is the total number of subjects) vectors were concatenated as rows of an $845 \times 4950$ matrix.
2) We aim to classify brain age values to two groups or classes. To do so, age values are converted to z-scores for all subjects. We then only keep subjects whose z-score values are above 0.5 or below -0.5. Thus, the subjects with age values in the left tail of the distribution are categorized as "low age subjects" with class '1', and those in the right tail of the distribution are designated as "high age subjects" with class '2'. In this way we discard those subjects with z-score values in the range $-0.5 < Z < 0.5$. The reason for discarding these subjects is to bring higher variance into the class distribution, and to better explain the brain analysis between these two groups. Doing so, we are left with $N = 568$ out of 845 subjects distributed in two classes. We apply these steps for all three sets of modalities.
3) Training and test data sets are normalized to zero-mean and unit variance. Each dataset is randomly divided into 90% training set and 10% test set. All models are run 50 times, and the mean and maximum values of the results are reported.
4) Results of ELM and proposed RES-ELM are compared with sparse Bayesian ELM (SB-ELM) [24], and randomly pruned ELM (RP-ELM). For RP-ELM, we randomly prune some features and classify data. The reason is that we would like to observe the effectiveness of the residual errors' criterion in detecting irrelevant features.
5) The setting for the sparsity parameter $S_p$ for RES-ELM and RP-ELM are $\{1, 2, …, 30\}$. So, if $S_p = 5$ this means that the learner keeps only 5% of training features for classification. The total number of hidden neurons for all ELM based NNs is in the range of $\{300, 325, 350, …, 2000\}$ with step size 25.
6) For all ELM based NNs, the regularization parameter $z$ is chosen in the range $\{10^{-8}, 10^{-4}, 10^{-2}, 10^{-1}, 1, 10, 10^2, 10^3\}$. For RES-ELM, this is applied to step 7 of Algorithm 1.
7) For RES-ELM, we also tune the threshold parameter $\varepsilon$ in (5) to be in the range $\{10^{-6}, 10^{-5}, …, 1, 5, 10, 50, 100\} \times 10^{-10}$.
8) For ELM based networks, the values of random weights $\boldsymbol{a}$ and bias $b$ for the sigmoid and RBF functions are drawn from the uniform distribution within $[-1, 1]$.
9) The proposed model is implemented on a personal computer with an Intel(R) processor, i5 core, 3.5 GHz, and 32 GB RAM. We use Matlab software to do the experiments.

### B. Results

Tables I and II show the test accuracies of ELM, SB-ELM, RP-ELM and the proposed RES-ELM using sigmoid and RBF hidden neurons, respectively. The results reveal that when we use more than one modality, we gain higher brain age classification accuracies. According to the results, RES-ELM has better test accuracy than all the counterparts and only SB-ELM is competitive with RES-ELM. The best overall average accuracy for all unimodal settings and multimodalities is achieved by RES-ELM.

TABLE I
TEST ACCURACY OF ELM BASED MODELS FOR SEVERAL DATASETS*
(SIGMOID NEURONS)

| Model/Data | ELM | | SB-ELM | | RP-ELM | | RES-ELM | |
|---|---|---|---|---|---|---|---|---|
| | Avg | Std | Avg | std | Avg | Std | Avg | Std |
| nb | 81.14 | 2.6 | 82.05 | 3.5 | 81.21 | 3.7 | **83.84** | 2.3 |
| rs | 80.88 | 3.0 | 81.79 | 3.3 | 81.55 | 3.8 | **83.33** | 2.3 |
| em | 78.13 | 2.4 | 82.47 | 3.4 | 81.11 | 3.7 | **82.85** | 2.7 |
| (nb, rs) | 81.15 | 3.0 | 83.75 | 3.8 | 82.06 | 3.6 | **85.26** | 2.5 |
| (nb, em) | 81.98 | 2.8 | 84.61 | 3.5 | 83.45 | 3.6 | **84.71** | 3.2 |
| (em, rs) | 80.25 | 3.1 | 82.02 | 3.5 | 81.17 | 4.0 | **84.53** | 2.7 |
| (nb, rs, em) | 81.08 | 3.1 | 81.02 | 3.8 | 82.06 | 3.8 | **83.05** | 2.5 |
| Average | 80.65 | 2.9 | 82.53 | 3.5 | 81.80 | 3.7 | **83.94** | 2.6 |

*Bold values indicate the best value under the same conditions.

TABLE II
TEST ACCURACY OF ELM BASED MODELS FOR SEVERAL DATASETS* (RBF NEURONS)

| Model/Data | ELM | | SB-ELM | | RP-ELM | | RES-ELM | |
|---|---|---|---|---|---|---|---|---|
| | Avg | Std | Avg | std | Avg | Std | Avg | Std |
| nb | 81.07 | 2.8 | 82.69 | 3.5 | 81.92 | 4.0 | **83.66** | 2.5 |
| rs | 80.29 | 3.1 | 81.55 | 3.4 | 81.07 | 4.0 | **83.17** | 2.6 |
| em | 78.47 | 2.6 | **82.81** | 3.4 | 81.54 | 3.9 | 82.75 | 2.8 |
| (nb, rs) | 80.38 | 3.2 | 83.46 | 3.2 | 82.16 | 4.2 | **85.45** | 3.0 |
| (nb, em) | 81.76 | 3.4 | 84.75 | 3.6 | 83.15 | 3.8 | **85.37** | 2.5 |
| (em, rs) | 81.78 | 3.2 | 81.84 | 3.5 | 81.17 | 4.2 | **84.88** | 2.7 |
| (nb, rs, em) | 81.38 | 3.3 | 82.68 | 3.4 | 80.09 | 4.0 | **82.83** | 2.8 |
| Average | 80.73 | 3.0 | 82.72 | 3.5 | 81.58 | 4.0 | **84.01** | 2.7 |

*Bold values indicate the best value under the same conditions.

It is worth noting that SB-ELM uses small fixed size of hidden neurons ranging from 20 to 210, which is the sparsest model. Comparing test accuracies for unimodal datasets with sigmoid activation function, RES-ELM gains higher average test accuracies (in %) 83.84, 83.33 and 82.85 for nb, rs and em fMRIs respectively against 82.05, 81.79 and 82.47 with SB-ELM as second best counterpart. For multimodal datasets, the gap of accuracies of RES-ELM compared to second best counterpart, SB-ELM is 1.51, 0.10, and 2.51 for bi-modalities {(nb, rs), (nb, es), (es, rs), (nb, rs, es)}-fMRIs respectively. If we compare overall average test accuracies of uni-modality and bi-modality, there is up to 2% gap in accuracies in favor of bi-modality datasets (with sigmoid neurons). When all modalities are concatenated, we do not observe a significant improvement compared to unimodal datasets and we observe a small reduction in accuracies of three-modality dataset compared with bi-modality datasets. One reason is the potential overfitting problem as adding an additional modality brings a higher ratio of the number of features to the number of instances.

The combination of (nb, rs) is the best choice among all modalities and gains highest accuracy for brain age prediction. We also observe that classical ELM has the poorest results among all learners because of the intense overfitting problem. Comparing the poorest (ELM) classifier and the best classifier (RES-ELM) for sigmoid function, the largest gap in overall average test accuracies is 4.7% for the em-fMRI and the smallest gap is 2% for all modality combination. Finally, the random RP-ELM has the highest gap between maximum and average accuracies. This means that the variance of results is high as we expect because features in RP-ELM are randomly selected. Similar interpretations go to RBF activation function for Table II.

Referring to Tables I and II, 'std' means standard deviation of learners for 50 iterations. RES-ELM tends to have the lowest standard deviation compared to the counterparts. This suggests that RES-ELM is not as affected by the random noise in the data as the other methods.

We observe a high variance of SB-ELM. The way for sparsity in SB-ELM is the same as the sparse Bayesian learning algorithm, called relevance vector machine (RVM) [31]. Due to the degeneracy of the covariance function in RVM, it has poor prediction capabilities if a test instance is distant from the relevance vectors [31]. Since the number of data instances for the fMRI modalities is small compared to the number of features, it is more likely that some test instances are far away from the relevance vectors. RP-ELM has the worst standard deviation among all classifiers since it chooses the features completely at random.

A statistical significance test based on a 95 percent confidence paired-t test is conducted for the comparison of two competitive models; i.e. RES-ELM and SB-ELM. The null hypothesis is rejected if the two compared means are unequal. Among all modalities, the null hypothesis is rejected for rs-fMRI, nb-fMRI and (nb, rs)-fMRI modalities. This means that the observed improvement in accuracy achieved by our proposed model over SB-ELM is statistically significant for these modalities for both Sigmoid and RBF hidden neurons under the performed test.

Fig. 4 illustrates a single run of ELM, RP-ELM and RES-ELM on all datasets. We choose the number of hidden neurons in the range {300, 400, ..., 1000} with step size 100. We notice that as the number of hidden neurons increases, the test accuracy of conventional ELM decreases. According to a theorem in ELM, as the number of training instances becomes equal to the number of features, the training accuracy tends to be optimal [14, 15]. So, this is a sign of an overfitting problem. Regarding sparsity-based ELM models, we observe that RES-ELM shows the least amount of perturbation as the number of hidden neurons increases. This is because adjusting the sparsity factor $Sp$ helps to reduce the overfitting problem. However, we do not observe any positive response in the performance of RP-ELM as the number of hidden neurons increases. The reason is that the selection of features in RP-ELM is by pure chance, unlike our approach in RES-ELM. While there is a potential to gain higher performance by discarding irrelevant features when the number of hidden neurons is high, RP-ELM randomly removes useful features too. That could explain why there is no gain in the performance of RP-ELM. The SB-ELM is the second best model in terms of classification accuracy.

Fig. 5 illustrates the restoration of residual errors of all competing methods versus steps 3 and 4 in the Algorithm 1, i.e., the error computation of $\mathbf{H}^{\dagger}\mathbf{H}$. SB-ELM has the lowest residual error of all the networks and RES-ELM has second rank while RP-ELM is unable to outperform RES-ELM since it prunes some neurons with low value residual errors. Note that the Y-axis is a log-scale.

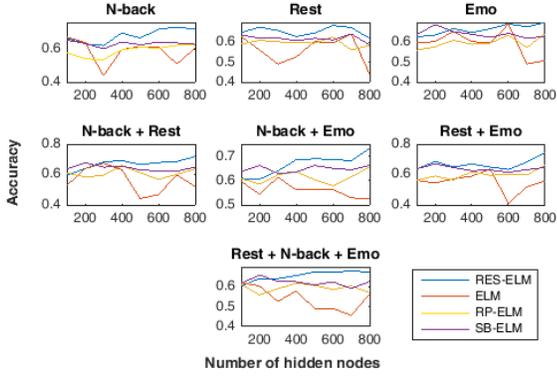

Fig. 4. Hidden neurons Vs classification accuracy (best view in color)

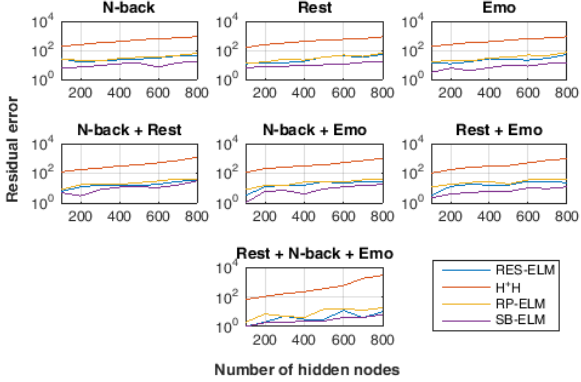

Fig. 5. Amount of residual errors before and after pruning (best view in color)

Tables III and IV show the number of hidden neurons used for unimodal and multimodal fMRI datasets with sigmoid and RBF neurons. SB-ELM uses lowest number of hidden neurons, because the range of neurons of SB-ELM is between 10 to 200 with step size 10. The training of SB-ELM is by far the slowest compared to all other considered methods, and we set this range of neurons to be the same as in [24]. However, according to Tables I and II, we observe that our proposed RES-ELM can gain higher performance than SB-ELM. Our proposed RES-ELM uses lower number of hidden neurons (see the column # used in Tables III and IV) for almost all modalities with both sigmoid and RBF neurons compared to ELM and RP-ELM.

TABLE III
NUMBER OF USED HIDDEN NEURONS FOR SEVERAL DATASETS (SIGMOID NEURONS)

| Dataset | ELM | SB-ELM | RP-ELM | RES-ELM |
|---|---|---|---|---|
| | # neur | (# neur, # used) | (# neur  # used) | (# neur, # used) |
| nb-fMRI | 375 | (200, **91**) | (1500, 150) | (1500, 150) |
| rs-fMRI | 425 | (190, **68**) | (1225, 122) | (1200, 120) |
| em-fMRI | 400 | (190, **74**) | (1525, 168) | (1625, 162) |
| (nb, rs)-fMRIs | 375 | (200, **90**) | (1375, 110) | (1425, 100) |
| (nb, em)-fMRIs | 400 | (200, **86**) | (1150, 115) | (1300, 117) |
| (em, rs)-fMRIs | 450 | (190, **83**) | (1500, 180) | (1775, 178) |
| (nb, rs, em)-fMRIs | 450 | (200, **102**) | (1825, 201) | (1475, 162) |
| Sum | 2875 | (1370, **594**) | (10100, 1046) | (10300, 989) |

*Bold values indicate the best value under the same conditions.

TABLE IV
NUMBER OF USED HIDDEN NEURONS FOR SEVERAL DATASETS (RBF NEURONS)

| Dataset | ELM | SB-ELM | RP-ELM | RES-ELM |
|---|---|---|---|---|
| | # neur | (# neur, # used) | (# neur  # used) | (# neur, # used) |
| nb-fMRI | 284 | (200, **87**) | (1525, 152) | (1425, 142) |
| rs-fMRI | 400 | (190, **78**) | (1200, 132) | (1200, 120) |
| em-fMRI | 375 | (190, **45**) | (1325, 172) | (1325, 132) |
| (nb, rs)-fMRIs | 375 | (200, **79**) | (1050, 194) | (1400, 126) |
| (nb, em)-fMRIs | 400 | (200, **99**) | (1575, 268) | (1150, 149) |
| (em, rs)-fMRIs | 475 | (190, **106**) | (1475, 265) | (1550, 232) |
| (nb, rs, em)-fMRIs | 525 | (200, **97**) | (1600, 272) | (1425, 228) |
| Average | 2834 | (1370, **591**) | (9784, 1455) | (9475, 1129) |

*Bold values indicate the best value under the same conditions.

In addition, it is interesting to look at the distribution of important features per modality. In this way we are able to recognize the importance of each modality in the prediction of the brain age. According to our findings, for (rs, es)-fMRI, 61% of important features belong to rs-fMRI. Comparing (nb, es)-fMRI, 56% of important features belong to nb-fMRI. Comparing (nb, rs)-fMRI, 55% belong to nb-fMRI. So, the nb-fMRI modality seems to be the most predictive of brain age, while the rs-fMRI modality is the second best. When we combine three modalities, the importance of modalities is unclear and each of them has almost the same contribution to the final brain age prediction.

## VI. CONCLUSION

In this study, a new sparse learning algorithm is proposed to learn informative features from functional connectivity (FC) measures derived from fMRI scans of adolescent brains using ELM based neural networks for the task of brain age prediction. Our proposed sparse learner takes advantage of the small errors overlooked during RSS minimization and hence a new feature learning is proposed. It can learn important features and integrate multimodal information. We found that the performance of brain age prediction can be further improved if combinations of FC information of different fMRI modalities are used.

## APPENDIX

*Proof.*

Suppose $\mathbf{H}_{L-1}^T \mathbf{H}_{L-1}$ has real and positive eigenvalues with descending order $\mu_1 > \mu_2 > \cdots > \mu_L$ and for $\mathbf{H}_L^T \mathbf{H}_L$ we also have, $\gamma_1 > \gamma_2 > \cdots > \gamma_L > \gamma_{L+1}$. The eigenvalues of $\mathbf{H}_{L-1}^T \mathbf{H}_{L-1}$ are interlaced with eigenvalues of $\mathbf{H}_L^T \mathbf{H}_L$ as it has been proven in the *Courant-Fischer Theorem* (See the example 7.5.3 of [32]):

$$\gamma_1 > \mu_1 > \gamma_2 > \mu_2 > \cdots > \gamma_L > \mu_L > \gamma_{L+1} \qquad (9)$$

From (5), the singular values in diagonal entries of **D**, $d_i$ are zero if $\delta_i \leq \varepsilon$, or equivalently if $\sqrt{\mu_i} \leq \varepsilon$ (we use the relation of eigenvalues and singular values, i.e. $\delta_i = \sqrt{\mu_i}$ [33]). In the following, we ignore the square root since it has no impact on our proof. The largest eigenvalue $\mu_1$ cannot be zero, since $\gamma_1$ is larger than the largest eigenvalue $\mu_1$, i.e., $\gamma_1 > \mu_1$, hence $\gamma_1$ cannot be zero by the definition of the residual error $E(\mathbf{H}_L) = \sum_{i:(\sqrt{\gamma_i}-\varepsilon) \leq 0} \sqrt{\frac{1}{\gamma_i}}$. Let's ignore $\gamma_1$ in (9) We then have,

$$\mu_1 > \gamma_2 > \mu_2 > \cdots > \gamma_L > \mu_L > \gamma_{L+1} \qquad (10)$$

or $\gamma_{i+1} < \mu_i$, $i = 1,2,\ldots,L$. This implies that when a new neuron is added, $\gamma_{i+1} < \mu_i$, or $\frac{1}{\gamma_{i+1}} > \frac{1}{\mu_i}$. So, it is expected that the error is larger than $E(\mathbf{H}_{L-1}) = \sum_{i:(\sqrt{\mu_i}-\varepsilon) \leq 0} \sqrt{\frac{1}{\mu_i}}$. Assume that $T$ eigenvalues of $\mathbf{H}_{L-1}^T \mathbf{H}_{L-1}$ ($T < L$) are set to zero, e.g., $\mu_i = 0$, $i = (L-T+1), (L-T+2), \ldots, L$, hence $\mathrm{E}(\mathbf{H}_{L-1}) = \sum_{i:(\sqrt{\mu_i}-\varepsilon) \leq 0} \sqrt{\frac{1}{\mu_i}}$, $i = L-T+1, \ldots, L$. Since $\gamma_{i+1} < \mu_i$, $i = 1, 2, \ldots, L$, the same procedure goes to $\gamma_{i+1} = 0$ for $i = (L-T+1), (L-T+2), \ldots, L$; i.e. $\mathrm{E}(\mathbf{H}_L) = \sum_{i:(\sqrt{\gamma_i}-\varepsilon) \leq 0} \sqrt{\frac{1}{\gamma_i}}$, $i = L-T+1, \ldots, L$.

The remaining $L$-$T$ eigenvalues $\mu_i$ in the iteration ($L$-$1$) are not less than the threshold $\varepsilon$; however, their new values $\gamma_{i+1}$ in the current iteration $L$ are less than their previous values $\mu_i$ in the iteration ($L$-$1$), i.e. $\gamma_{i+1} < \mu_i$, $i = 1, \ldots, L-T$. Since the new values become smaller, 1) if at least one of them is less than the threshold $\varepsilon$ we conclude $\mathrm{E}(\mathrm{H}_L) > \mathrm{E}(\mathrm{H}_{L-1})$ and, 2) If all of them are not less than the threshold $\varepsilon$ in (5), we conclude $\mathrm{E}(\mathrm{H}_L) = \mathrm{E}(\mathrm{H}_{L-1})$. Thus, $\mathrm{E}(\mathrm{H}_L) \geq \mathrm{E}(\mathrm{H}_{L-1})$ and the proof for (7) is concluded.